\documentclass[sigconf]{acmart}

\setcopyright{none}
\usepackage{placeins}
\usepackage{adjustbox}
\usepackage{tikz}
\usepackage{float}

\usetikzlibrary{positioning,arrows.meta,calc}
\renewcommand\footnotetextcopyrightpermission[1]{}

\begin{document}


\title{Evaluating LLMs as Judges of Counterspeech Quality Against Online Hate Speech}

\acmConference[]{Preprint}{Authors' Version}{2026}
\acmYear{}

\author{Johnny Alali}
\email{alalijohnny@gmail.com}
\affiliation{%
  \institution{Vrije Universiteit Amsterdam}
  \city{Amsterdam}
  \country{The Netherlands}
}

\author{Nicolás Benjamín Ocampo}
\email{n.b.ocampo@cwi.nl}
\affiliation{%
  \institution{Centrum Wiskunde \& Informatica}
  \city{Amsterdam}
  \country{The Netherlands}
}

\author{Davide Ceolin}
\email{davide.ceolin@cwi.nl}
\affiliation{%
  \institution{Centrum Wiskunde \& Informatica}
  \city{Amsterdam}
  \country{The Netherlands}
}

\renewcommand{\shortauthors}{Alali et al.}


\begin{abstract}

Prolonged exposure to hate speech on social media can cause lasting psychological distress among users. Counterspeech has emerged as an effective, non-censorial mitigation strategy, but its deployment requires rigorous quality assessment. Manual assessment is unscalable due to the large volume of content on the Web and imposes severe emotional strain on human annotators. Large Language Models (LLMs) offer a scalable alternative, yet concerns remain regarding their reliability. In this paper, we investigate the extent to which LLMs can effectively perform this task. We test 12 open-weight LLMs of different sizes and families under zero-shot, few-shot, and chain-of-thought prompting, to approximate pointwise and pairwise human expert assessment of counterspeech quality across four dimensions: relevance, coherence, aggressiveness, and suitableness.
We find that (i) LLMs show low agreement and moderate rank correlations with human experts: our best configuration, Qwen 2.5 72B on coherence, achieves up to a linear-weighted Cohen's $\kappa_l = 0.19$ and Spearman's $\rho = 0.42$ in pointwise evaluation and an unweighted Cohen's $\kappa = 0.25$ in pairwise evaluation. (ii) Converting pointwise scores into pairwise labels substantially improves human--LLM agreement for selected configurations. For same-instance coherence, Qwen 2.5 72B under zero-shot prompting achieves $\kappa = 0.35$ with converted labels, compared with $\kappa = 0.25$ when judging pairs directly, and Mixtral 8$\times$7B under few-shot prompting improves from $\kappa = 0.07$ to $\kappa = 0.34$. (iii) LLMs agree more with one another than with human experts, particularly in pairwise evaluation, where the strongest model pair, OLMo 2 32B and Llama 3.3 70B on coherence, reaches substantial agreement with $\kappa = 0.76$, whereas their agreement with the experts in the same configuration does not exceed $\kappa = 0.21$. 

\end{abstract}



\maketitle


\section{Introduction}
\label{sec:intro}

Online platforms have become a major channel for the dissemination of hate
speech, which has been linked to offline violence~\cite{muller2021fanning} and
to psychological distress among those exposed to it~\cite{saha2019prevalence}.
Counterspeech has emerged as an effective mitigation strategy to oppose online
hate. Rather than suppressing hateful content, it challenges it through a
non-censorial response~\cite{benesch2014countering,mathew2019thou,
chung2024understanding}. Evidence of its effectiveness was shown on Twitter
users who received empathic counterspeech as a response to xenophobic posts and
were less likely to produce hateful content over a four-week
period~\cite{hangartner2021empathy}.

However, deploying counterspeech requires rigorous human assessment, involving several
factors such as evaluation criteria, rating scales, quality dimensions, and
annotator expertise~\cite{bonaldi2024nlp}. In this context, two main evaluation
approaches have been adopted: Likert or continuous-scale ratings of hate speech
and counterspeech message pairs~\cite{chung2021towards,tekiroglu2022using,
jones2024multi,bonaldi2024safer,hengle2025cseval}, and pairwise comparisons, in
which the better of two counterspeech responses to a given hate speech message
is selected~\cite{tekiroglu2022using,zubiaga2024llm}.
These approaches offer different advantages and limitations. Likert scales are
convenient because they assign a direct score to each counterspeech response,
but annotators may struggle to distinguish between intermediate
values~\cite{kiritchenko2017bws}, and the resulting ordinal ratings are often
treated as interval-scale measurements~\cite{liddell2018analyzing}. In contrast,
pairwise comparisons simplify the task for human annotators by requiring them to
choose between two responses, but they come at the cost of requiring a larger
number of comparisons~\cite{vanderlee2019best,zubiaga2024llm}. With either
approach, manual assessment is difficult to scale to the volume of content on
the Web and repeatedly exposes annotators to hateful content.

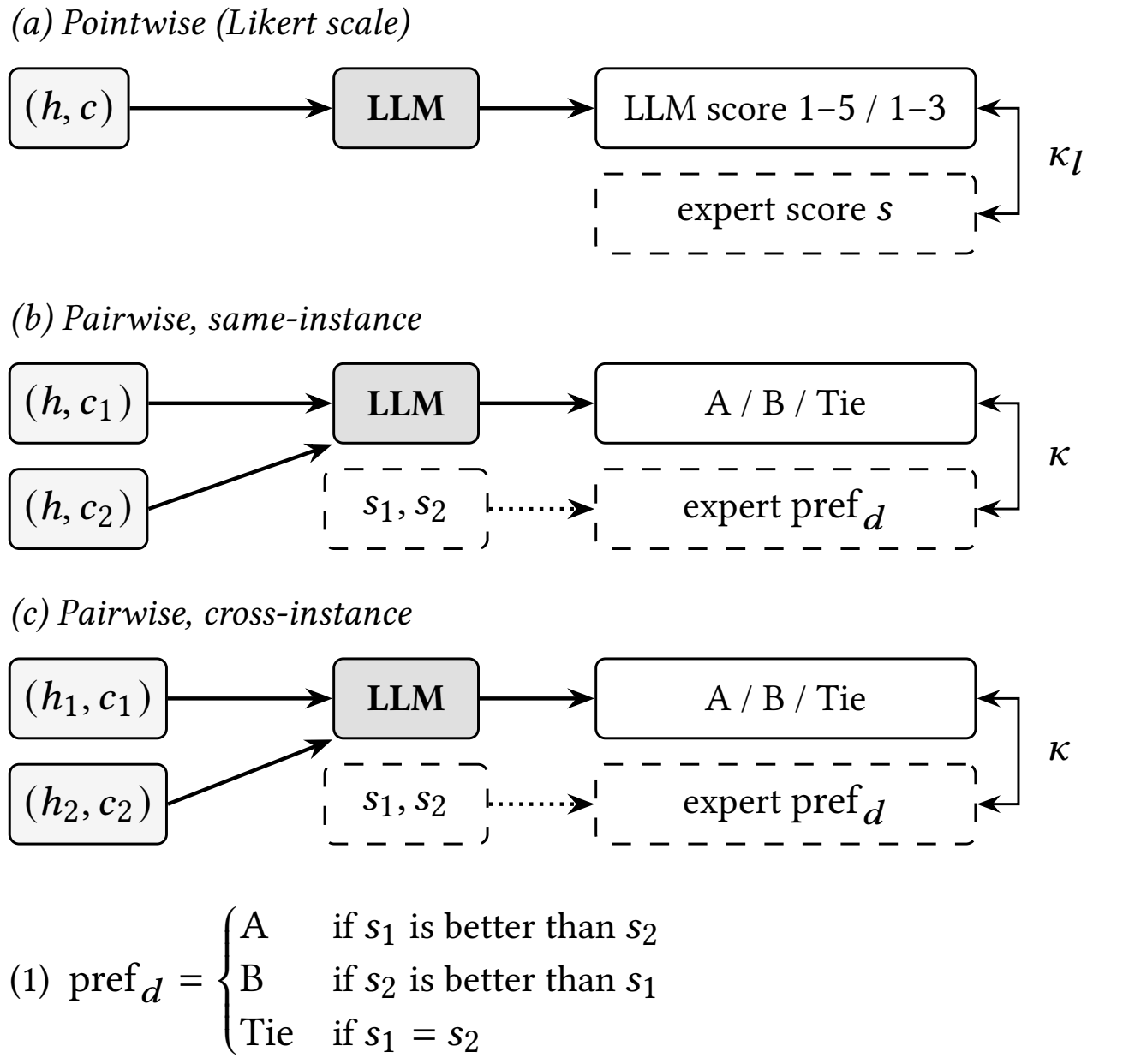
\begin{figure}[t]
\centering
\resizebox{\columnwidth}{!}{%
\begin{tikzpicture}[
  font=\scriptsize,
  box/.style={draw, rounded corners=1.5pt, minimum height=4.4mm, inner sep=1.8pt},
  inp/.style={box, fill=black!4},
  llm/.style={box, minimum width=8mm, fill=black!12, font=\scriptsize\bfseries},
  outp/.style={box, minimum width=21mm},
  ref/.style={box, dashed, minimum width=21mm},
  sc/.style={box, dashed, minimum width=9mm},
  arr/.style={-{Stealth[length=1.5mm]}, semithick},
  cmp/.style={{Stealth[length=1.3mm]}-{Stealth[length=1.3mm]}},
  lab/.style={font=\scriptsize\itshape, anchor=north west, inner sep=0pt},
]
\def\xllm{22mm}
\def\xout{43mm}

\node[lab] (la) at (0,0) {(a) Pointwise (Likert scale)};
\node[inp, anchor=north west] (pin)
  at ($(la.south west)+(0,-1.3mm)$) {$(h, c)$};
\node[llm] (pl) at (\xllm, 0 |- pin) {LLM};
\node[outp] (pout) at (\xout, 0 |- pin) {LLM score $1$--$5$ / $1$--$3$};
\node[ref, below=1.3mm of pout] (pref) {expert score $s$};
\draw[arr] (pin) -- (pl);
\draw[arr] (pl) -- (pout);
\draw[cmp] (pout.east) -- ++(2.3mm,0)
  |- node[pos=0.25, right=0.5mm] {$\kappa_l$} (pref.east);

\node[lab] (lb) at ($(la.west |- pref.south)+(0,-2.6mm)$)
  {(b) Pairwise, same-instance};
\node[inp, anchor=north west] (b1)
  at ($(lb.south west)+(0,-1.3mm)$) {$(h, c_1)$};
\node[inp, below=1.3mm of b1] (b2) {$(h, c_2)$};
\node[llm] (bl) at (\xllm, 0 |- b1) {LLM};
\node[outp] (bout) at (\xout, 0 |- b1) {A / B / Tie};
\node[sc] (bs) at (\xllm, 0 |- b2) {$s_1, s_2$};
\node[ref] (bref) at (\xout, 0 |- b2) {expert $\mathrm{pref}_d$};
\draw[arr] (b1) -- (bl);
\draw[arr] (b2.east) -- (bl.south west);
\draw[arr] (bl) -- (bout);
\draw[arr, densely dotted] (bs) -- (bref);
\draw[cmp] (bout.east) -- ++(2.3mm,0)
  |- node[pos=0.25, right=0.5mm] {$\kappa$} (bref.east);

\node[lab] (lc) at ($(la.west |- bref.south)+(0,-2.6mm)$)
  {(c) Pairwise, cross-instance};
\node[inp, anchor=north west] (c1)
  at ($(lc.south west)+(0,-1.3mm)$) {$(h_1, c_1)$};
\node[inp, below=1.3mm of c1] (c2) {$(h_2, c_2)$};
\node[llm] (cl) at (\xllm, 0 |- c1) {LLM};
\node[outp] (cout) at (\xout, 0 |- c1) {A / B / Tie};
\node[sc] (cs) at (\xllm, 0 |- c2) {$s_1, s_2$};
\node[ref] (cref) at (\xout, 0 |- c2) {expert $\mathrm{pref}_d$};
\draw[arr] (c1) -- (cl);
\draw[arr] (c2.east) -- (cl.south west);
\draw[arr] (cl) -- (cout);
\draw[arr, densely dotted] (cs) -- (cref);
\draw[cmp] (cout.east) -- ++(2.3mm,0)
  |- node[pos=0.25, right=0.5mm] {$\kappa$} (cref.east);

\node[anchor=north west, inner sep=0pt] (eq)
  at ($(la.west |- c2.south)+(0,-3mm)$)
  {(1)\enspace
   $\mathrm{pref}_d =
   \begin{cases}
     \mathrm{A}   & \text{if } s_1 \text{ is better than } s_2\\
     \mathrm{B}   & \text{if } s_2 \text{ is better than } s_1\\
     \mathrm{Tie} & \text{if } s_1 = s_2
   \end{cases}$};

\end{tikzpicture}%
}
\caption{Overview of the evaluation settings. In setting (a), a single counterspeech item for one hateful instance is evaluated on a Likert scale. In settings (b) and (c), two items are compared: in (b) on the same hateful instance, while in (c) on two different instances matched by target group and counterspeech strategy.}
\label{fig:paradigms}
\end{figure}
\FloatBarrier

Large language models (LLMs) have emerged as a scalable alternative, including
fine-tuned evaluators such as JudgeLM~\cite{zhu2025judgelm}, used in the shared
task of the First Workshop on Multilingual Counterspeech
Generation~\cite{bonaldi2025mcg}, and ArgJudge~\cite{lin2023argue}, which
scores the argumentative quality of counter-arguments. However, the conditions
under which these models perform reliably remain unclear, and findings regarding
their capabilities are contrasting: while LLM evaluators correlate more strongly
with human judgments of counter-narratives than reference-based
metrics~\cite{jones2024multi,zubiaga2024llm}, their agreement with human
judgments varies widely across models, datasets, and evaluated
properties~\cite{bavaresco2025llms}, and open-weight models correlate
substantially less with expert judgments than GPT-4~\cite{hengle2025cseval}.
Moreover, a systematic comparison of LLMs across pointwise and pairwise
evaluation setups remains lacking.

This study uses CSEval~\cite{hengle2025cseval} as a benchmark to evaluate
LLMs on counterspeech quality assessment, comparing their judgments with
the expert human annotations it provides. Since counterspeech quality is
multidimensional, we examine each quality dimension separately. We aim to
address the following research question:
\par\vspace{5pt}
\noindent\hspace{1.5em}%
\begin{minipage}{\dimexpr\columnwidth-3em\relax}
\emph{How well can LLMs reproduce expert evaluation of counterspeech quality
against hate speech on social media?}
\end{minipage}
\par\vspace{5pt}

\noindent Our contribution is an extensive evaluation of 12 open-weight LLMs from six model families, ranging in size from 7B to 141B, under zero-shot, few-shot, and chain-of-thought prompting to assess counterspeech quality across four dimensions: relevance, coherence, aggressiveness, and suitableness. We evaluate LLMs in both pointwise and pairwise settings (Figure~\ref{fig:paradigms}). For pairwise evaluation, we consider two settings: same-instance comparisons, which compare counterspeech responses to the same hate speech message, and cross-instance comparisons, which compare responses to different hate speech messages sharing the same target group and counterspeech
strategy.
Leveraging this contribution, we derive the following
insights: (i) LLMs show low agreement and moderate rank correlations with human experts: our best configuration, Qwen 2.5 72B on coherence, achieves up to a linear-weighted Cohen's $\kappa_l = 0.19$ and Spearman's $\rho =0.42$ in pointwise evaluation and an unweighted Cohen's $\kappa = 0.25$ in pairwise evaluation. 
(ii) Converting pointwise scores into pairwise labels substantially improves human--LLM agreement for selected configurations. For same-instance coherence, Qwen 2.5 72B under zero-shot prompting achieves $\kappa = 0.35$ with converted labels, compared with $\kappa = 0.25$ when judging pairs directly, and Mixtral 8$\times$7B under few-shot prompting improves from $\kappa = 0.07$ to $\kappa = 0.34$. (iii) LLMs agree more with one another than with human experts, particularly in pairwise evaluation, where the strongest model pair, OLMo 2 32B and Llama 3.3 70B on coherence, reaches substantial agreement with $\kappa = 0.76$, whereas their agreement with the experts in the same configuration does not exceed $\kappa = 0.21$.

The remainder of this paper is organized as follows.
Section~\ref{sec:related-work} reviews related work on counterspeech evaluation
and LLMs as judges. Section~\ref{sec:methodology} describes the pointwise and
pairwise evaluation settings, Section~\ref{sec:methodology-data} the dataset,
and Section~\ref{sec:methodology-judges} the models, prompting strategies, and
agreement measures. Section~\ref{sec:results_discussion} presents and discusses
the results, and Section~\ref{sec:conclusion} concludes.


\section{Related Work}
\label{sec:related-work}

Counterspeech offers a constructive alternative to restrictive
approaches for addressing hate speech on social media. Assessing
counterspeech quality is important for effective intervention, but human evaluation can become a bottleneck at scale. We review existing approaches to intrinsic
counterspeech evaluation, which assess generated responses independently
of their downstream effects, ranging from automatic metrics to human
and LLM-based judgments.

\subsection{Metrics for Counterspeech Evaluation}
\label{sec:relatedwork-counterspeech}

Reference-based metrics such as BLEU~\cite{papineni2002bleu},
ROUGE~\cite{lin2004rouge}, METEOR~\cite{banerjee2005meteor}, and
BERTScore~\cite{zhang2020bertscore} assess generated counterspeech
through lexical or semantic similarity to reference texts. However,
multiple valid responses may exist for the same hate speech
message~\cite{hengle2025cseval}, and reference similarity does not
necessarily capture specific quality dimensions~\cite{bonaldi2024nlp}.
Other automatic metrics assess individual properties rather than
overall quality, including repetitiveness, measured with the repetition
rate~\cite{bertoldi2013cache,cettolo2014repetition} as in counterspeech
generation studies~\cite{chung2021towards,tekiroglu2022using},
safety~\cite{markov2023holistic,bonaldi2024safer}, and knowledge
overlap~\cite{chung2021towards}.

Human evaluation often assesses counterspeech pointwise using
Likert scales~\cite{jones2024multi,cima2025contextualized}
across dimensions such as
coherence~\cite{chung2021towards,hengle2025cseval},
grammaticality~\cite{tekiroglu2022using},
aggressiveness~\cite{hengle2025cseval},
informativeness~\cite{chung2021towards,bonaldi2024safer},
cogency~\cite{bonaldi2024safer}, and
suitableness~\cite{tekiroglu2022using,hengle2025cseval}. However, distinguishing intermediate scores
can be challenging~\cite{kiritchenko2017bws}. Pairwise evaluation
instead selects the preferred response from two alternatives~\cite{zubiaga2024llm}. While this simplifies the judgment, it does not directly quantify the magnitude of quality
differences, and exhaustive comparisons grow quadratically with
the number of systems compared~\cite{vanderlee2019best,zubiaga2024llm}.
Hybrid approaches combine pointwise ratings with comparative
selection among responses to the same hate speech
message~\cite{tekiroglu2022using}.

Overall, counterspeech evaluation remains methodologically
heterogeneous, with studies combining reference-based,
property-based, and human evaluations~\cite{chung2021towards,
tekiroglu2022using,bonaldi2024nlp}. The reliance on human
judgments also presents scalability challenges, motivating
the investigation of automated evaluation methods.

\subsection{LLMs as Evaluators}
\label{sec:relatedwork-llmevaluators}

LLM-based evaluation offers a potential alternative to human assessment
of counterspeech quality, although evidence of its reliability remains
mixed. Several studies report encouraging results for evaluating
responses to hate speech. LLM evaluators achieve stronger correlations
with human judgments than reference-based metrics in counter-narrative
evaluation~\cite{jones2024multi}, and an LLM that compares the outputs of different counter-narrative generation systems in pairs produces system rankings that correlate strongly with human preferences~\cite{zubiaga2024llm}. Auto-CSEval, which calibrates evaluation instructions using expert judgments, improves correlation with human ratings of counterspeech
over conventional metrics and LLM-based baselines~\cite{hengle2025cseval}.

These studies nevertheless leave open how closely LLM judgments agree
with human judgments. The strong correlation reported by Zubiaga et al.
concerns rankings of generation systems rather than agreement on
individual responses to hate speech~\cite{zubiaga2024llm}, and on CSEval,
the open-weight models evaluated by Hengle et al. achieve substantially
lower correlations with expert judgments than GPT-4~\cite{hengle2025cseval}. Findings on prompting vary as
well: chain-of-thought (CoT)-based evaluation improves correlation with expert
judgments across all four dimensions for GPT-4 on CSEval, while improvements are
less consistent across dimensions for the open-weight models
examined~\cite{hengle2025cseval}. LLMs may also suffer from position
bias in pairwise evaluation~\cite{zheng2023judging}. Together, these
findings highlight the importance of validating LLM evaluators against human
judgments for each model, prompting strategy, evaluation setting, and quality
dimension. The LLM-based counterspeech evaluation studies discussed above
primarily report correlations with human judgments rather than
chance-corrected agreement measures such as Cohen's $\kappa$,
and focus on specific evaluation settings. Our work builds on CSEval's expert annotations and addresses these
gaps in three ways: First, we examine both pointwise and pairwise evaluation
of counterspeech responses, including same-instance and cross-instance
comparisons, and compare pairwise judgments with labels converted from pointwise
scores. Second, we evaluate 12 open-weight LLMs from six model families and of
different sizes under zero-shot, few-shot, and chain-of-thought prompting, and
measure agreement with the experts using Cohen's $\kappa$, which corrects for
chance agreement. Third, we compare LLM--LLM agreement with human--LLM agreement
to examine whether agreement among LLMs corresponds to agreement with
experts.


\section{Methodology}
\label{sec:methodology}
\label{sec:methodology-pointwise}
\label{sec:methodology-pairwise}

We test LLMs as counterspeech evaluators under pointwise and pairwise evaluation, varying the model, the prompting strategy, and the quality dimension, and compare their judgments with expert human annotations.

Figure~\ref{fig:paradigms} illustrates the two evaluation settings: pointwise evaluation (Figure~\ref{fig:paradigms}a) and pairwise evaluation with same-instance and cross-instance comparisons (Figure~\ref{fig:paradigms}b--c). In the pointwise setting, the model receives a hate speech message $h$ and a single counterspeech response $c$, and assigns an integer score $s$ on the rating scale of the evaluated quality dimension. Each quality dimension is evaluated independently, and the model's scores are compared with the expert scores for the same instances. In the pairwise setting, the model receives two counterspeech responses $c_1$ and $c_2$, presented as Counterspeech~A and Counterspeech~B, and judges which of the two is better on the evaluated quality dimension. It returns one of three labels: \texttt{A} or \texttt{B} if one response is better, or \texttt{Tie} if neither response is clearly better than the other. We consider two types of comparisons: in same-instance comparisons, counterspeech $c_1$ and $c_2$ address the same hate speech message $h$, whereas in cross-instance comparisons, $h_1$ and $h_2$ are distinct hate speech messages targeting the same group, addressed by counterspeech $c_1$ and $c_2$, respectively.

Pointwise scores on a rating scale can be converted into pairwise labels. Given the scores $s_1$ and $s_2$ of two responses on dimension $d$, the counterspeech with the better score is preferred, and equal scores yield a tie (Eq.~(1) in Figure~\ref{fig:paradigms}). We use this rule to convert each model's pointwise scores into pairwise labels for the same response pairs judged in the pairwise setting. Both settings then share the \{\texttt{A}, \texttt{B}, \texttt{Tie}\} classes, so pointwise and pairwise evaluation can be compared directly on the same pairs. The same rule, applied to expert scores, provides the reference labels for evaluating both pointwise-derived and direct pairwise judgments.


\section{Dataset}
\label{sec:methodology-data}

We use CSEval~\cite{hengle2025cseval}, a benchmark for multidimensional counterspeech evaluation annotated by domain experts. CSEval was built by sampling hate speech messages from IntentCONAN~\cite{gupta2023counterspeeches}, a dataset in which each message has several valid counterspeech responses, and contains 2,223 unique hate speech messages, 4,318 human-written reference counterspeech responses, and 7,926 responses generated by models~\cite{hengle2025cseval}. Five expert annotators with research backgrounds in hate speech and counterspeech scored hate speech--counterspeech pairs along four quality dimensions: \emph{relevance}, which measures how well the counterspeech addresses the hate speech; \emph{coherence}, which captures whether it provides specific and coherent arguments that counter the hate speech; \emph{aggressiveness}, which measures confrontational or inflammatory language; and \emph{suitableness}, which assesses whether the response could be used as counterspeech without editing. Relevance, coherence, and aggressiveness are scored on 1--5 Likert scales, whereas suitableness uses a 1--3 scale, since it is close to a yes/no judgment of whether a response can be used as is~\cite{tekiroglu2022using}. Higher scores indicate better quality for all dimensions except aggressiveness, for which lower scores are preferred, as less aggressive counterspeech is considered more effective~\cite{hengle2025cseval}.

CSEval contains expert scores for two types of counterspeech: human-written reference responses, which form its gold-standard subset, and responses generated by five systems~\cite{hengle2025cseval}. We evaluate only human-written counterspeech: the human-written reference responses form our evaluation set, and their expert scores serve as the expert reference. The model-generated responses and their expert scores are used only to select few-shot demonstrations and are never evaluated. We draw demonstrations from this subset for two reasons: it keeps the demonstrations disjoint from the evaluated instances, avoiding any overlap between examples and test items, and it covers the full range of scores, including clearly poor responses that illustrate the lower end of each scale. Because these demonstrations are model-generated, they may differ in style from the human-written responses being evaluated, which could affect few-shot performance.

CSEval marks missing annotations with a value of 0, outside the defined rating scales; we exclude these values and remove instances without any valid annotation. The pointwise evaluation uses 1,400 hate speech--counterspeech instances spanning 1,056 unique hate speech messages. Because some instances lack an annotation for individual dimensions, the number of evaluable instances varies by dimension. Table~\ref{tab:data} reports the number of evaluable instances and the distribution of expert scores for each dimension.

\begin{table}[H]
\caption{Expert score distribution in the pointwise evaluation set (1,400
instances). Suitableness uses a 1--3 scale.}
\label{tab:data}
\begin{tabular}{lrrrrrr}
\toprule
Dimension & $n$ & 1 & 2 & 3 & 4 & 5 \\
\midrule
Relevance      & 1,389 & 296 & 811 & 217 & 65  & 0  \\
Coherence      & 1,397 & 612 & 712 & 57  & 16  & 0  \\
Aggressiveness & 1,092 & 91  & 225 & 568 & 194 & 14 \\
Suitableness   & 1,386 & 693 & 542 & 151 & --  & -- \\
\bottomrule
\end{tabular}
\end{table}
\FloatBarrier

Since CSEval provides no pairwise annotations, we construct pairwise datasets from these pointwise instances and derive reference labels for each pair from the expert scores of its two responses using the rule in Eq.~(1) (Section~\ref{sec:methodology}): the response with the better expert score is preferred, and equal scores yield a \texttt{Tie}. The pairwise datasets are built from 1,398 instances: two of the 1,400 instances have exactly the same text as another instance and differ only in their identifier, so we remove them to avoid pairing a response with itself. The \emph{same-instance} dataset compares alternative counterspeech responses addressing the same hate speech message: 269 of the 1,056 unique hate speech messages have between two and ten responses, and pairing these responses yields 454 comparisons. The \emph{cross-instance} dataset compares responses addressing different hate speech messages. To construct comparable pairs, we pair instances only when they share the same target group and counterspeech strategy, which yields 7,047 comparisons.

Target-group labels were obtained by matching each hate speech message to its source dataset: 964 instances were matched to IntentCONANv2~\cite{hengle2024intent}, which pairs hate speech with counterspeech written for different intents (e.g., informative, questioning, denouncing, positive), and 323 to MultiCONAN~\cite{fanton2021human}, a multi-target dataset of hate speech--counterspeech pairs collected with a human-in-the-loop procedure. The remaining 113 instances, which appear in neither dataset, were assigned to a target group used in the source datasets by one of the authors, based on the content of each hate speech message; these labels were not checked by a second annotator. The instances cover 15 target groups. The six largest are Muslims (463 instances), women (226), Black people (177), LGBTQ+ people (152), immigrants (135), and Jews (118); the remaining nine, including Asian people, refugees, and people with disabilities, each have fewer than 25 instances and are combined as Other (129 instances).

If either response has no score for the evaluated dimension, the comparison is excluded for that dimension. Table~\ref{tab:pairs} reports the resulting number of comparisons per dimension and the distribution of reference labels.

\begin{table}[H]
\caption{Number of pairwise comparisons per dimension and distribution of reference labels.}
\label{tab:pairs}
\small
\setlength{\tabcolsep}{3.5pt}
\begin{tabular}{lrrrr|rrrr}
\toprule
 & \multicolumn{4}{c|}{Same-instance} & \multicolumn{4}{c}{Cross-instance} \\
Dimension & $n$ & A & B & Tie & $n$ & A & B & Tie \\
\midrule
Relevance      & 441 & 104 & 106 & 231 & 6,978 & 1,884 & 2,007 & 3,087 \\
Coherence      & 450 & 115 & 111 & 224 & 7,047 & 1,626 & 1,703 & 3,718 \\
Aggressiveness & 263 & 68  & 66  & 129 & 5,963 & 1,890 & 1,826 & 2,247 \\
Suitableness   & 447 & 102 & 104 & 241 & 6,890 & 2,050 & 1,850 & 2,990 \\
\bottomrule
\end{tabular}
\end{table}


\section{Experimental Setup}
\label{sec:methodology-judges}

We evaluate twelve open-weight LLMs from six model families:
Llama~\cite{llama2024herd},
OLMo 2~\cite{olmo2025furious},
Qwen 2.5~\cite{qwen2024qwen25},
Mistral~\cite{jiang2023mistral,mistral2024small},
Mixtral~\cite{jiang2024mixtral}, and
Command R~\cite{cohere2024commandr7b,cohere2024commandrplus}.
We have excluded closed models due to replicability concerns.
For each family, we include a smaller and a larger variant to examine
whether model scale is associated with agreement with human judgments.
For analysis, we group models into small (7--8B), medium (13--39B active
parameters), and large (70--104B) models. Table~\ref{tab:models} lists the model
identifiers, parameter counts, and size groups.

\begin{table}[!htbp]
\centering
\caption{Evaluated open-weight LLMs. For the Mixtral mixture-of-experts models,
size groups are based on active parameters.}
\label{tab:models}
\scriptsize
\begin{tabular}{llrll}
\toprule
Family & Model & Size & Group & Identifier \\
\midrule
Llama & Llama 3.1 8B & 8B & Small & Llama-3.1-8B-Instruct \\
 & Llama 3.3 70B & 70B & Large & Llama-3.3-70B-Instruct \\
OLMo 2 & OLMo 2 7B & 7B & Small & OLMo-2-1124-7B-Instruct \\
 & OLMo 2 32B & 32B & Medium & OLMo-2-0325-32B-Instruct \\
Qwen 2.5 & Qwen 2.5 7B & 7B & Small & Qwen2.5-7B-Instruct \\
 & Qwen 2.5 72B & 72B & Large & Qwen2.5-72B-Instruct \\
Mistral & Mistral 7B & 7B & Small & Mistral-7B-Instruct-v0.3 \\
 & Mistral 22B & 22B & Medium & Mistral-Small-Instruct-2409 \\
Mixtral & Mixtral 8$\times$7B & 47B & Medium & Mixtral-8x7B-Instruct-v0.1 \\
 & Mixtral 8$\times$22B & 141B & Medium & Mixtral-8x22B-Instruct-v0.1 \\
Command R & Command R 7B & 7B & Small & Command-R7B-12-2024 \\
 & Command R+ 104B & 104B & Large & Command-R+-08-2024 \\
\bottomrule
\end{tabular}
\end{table}

We evaluate each model under zero-shot, few-shot, and chain-of-thought (CoT) prompting. Zero-shot 
uses the CSEval definition of the target quality dimension without demonstrations. Few-shot 
adds target-group-specific demonstrations. In pointwise evaluation, five demonstrations cover the 1--5 scale for relevance, coherence, and aggressiveness, while three cover the 1--3 suitableness scale. In pairwise evaluation, three demonstrations illustrate the three possible outcomes: \texttt{A}, \texttt{B}, and \texttt{Tie}. 
CoT adds predefined dimension-specific criteria that decompose each dimension into a sequence of evaluation questions. We used a temperature of 0 following Bavaresco et al.~\cite{bavaresco2025llms}, which reduces sampling randomness. Outputs are constrained to the valid score or pairwise labels, so the model returns only the final score or label and no reasoning trace is generated.

For pointwise evaluation, our primary agreement measure is linear-weighted
Cohen's $\kappa$~\cite{cohen1968weighted}, which accounts for chance agreement
while weighting disagreements by their ordinal distance. We additionally report
Spearman's rank correlation ($\rho$).

For pairwise evaluation, pairwise labels are treated as nominal categories, so we
use unweighted Cohen's $\kappa$~\cite{cohen1960coefficient} against the reference labels. For the converted labels derived from each model's pointwise scores, we compute the same unweighted $\kappa$ against the reference labels, denoted $\kappa_{\mathrm{pw}}$, so that pointwise and pairwise evaluation can be compared on the same pairs. LLM--LLM agreement is computed with
the same measures between two models. We interpret $\kappa$ values following Landis and Koch~\cite{landis1977measurement}, who label agreement as slight for $\kappa \le 0.20$, fair for $0.21$--$0.40$, moderate for $0.41$--$0.60$, substantial for $0.61$--$0.80$, and almost perfect above $0.80$.

As reference points, we report a majority baseline, which always predicts the
most frequent expert score or label and therefore yields $\kappa = 0$: since it
always predicts the same label, its observed agreement equals the proportion of
that label among the expert annotations, which is exactly the agreement expected
by chance. We also report a random baseline, which predicts scores or labels
uniformly at random. Since random predictions yield $\kappa$ values scattered
around zero, we report the upper bound of the 95\% range of $\kappa$ over 2,000
random draws: a configuration that exceeds this value is unlikely to reflect
chance agreement.


\section{Results and Discussion}
\label{sec:results_discussion}

We test three comparisons between configurations with non-parametric tests on
paired agreement values: (i) zero-shot vs.\ few-shot vs.\ chain-of-thought
prompting, paired by model and quality dimension, using the Friedman test, the
non-parametric equivalent of a repeated-measures analysis of variance for three
or more related conditions, followed by Wilcoxon signed-rank tests of few-shot
vs.\ zero-shot and chain-of-thought vs.\ zero-shot prompting; (ii) the larger
vs.\ the smaller model of each family, paired by prompting strategy and quality
dimension, using the Wilcoxon signed-rank test; and (iii) direct pairwise
judgments vs.\ labels converted from pointwise scores, paired by model,
prompting strategy, and quality dimension, using the Wilcoxon signed-rank test.
Confidence intervals are obtained with a cluster bootstrap that resamples hate
speech messages together with all their responses (10,000 resamples). For a
single configuration, we compare $\kappa$ and $\kappa_{\mathrm{pw}}$ with the
same bootstrap, applied to their difference. Per-family comparisons are
adjusted for multiple testing with the Benjamini--Hochberg procedure
(Appendix~\ref{app:stat-tests}).

\subsection{Human--LLM Agreement in Pointwise Evaluation}
\label{sec:results-pointwise}

Table~\ref{tab:pointwise-full} reports agreement for all 144 pointwise
configurations, covering 12 models, three prompting strategies, and four quality
dimensions.

No pointwise configuration exceeds slight agreement on the
Landis--Koch scale ($\kappa_l \le 0.20$), although rank correlations
are moderate for some dimensions. The highest agreement is obtained
by Qwen 2.5 72B for coherence under few-shot prompting, with
$\kappa_l = 0.191$, 95\% CI [0.163, 0.219], followed by
OLMo 2 32B for coherence under CoT prompting, with $\kappa_l = 0.188$
[0.162, 0.213]. Spearman correlations reach $\rho = 0.431$, obtained by
Mixtral 8$\times$22B for aggressiveness under zero-shot prompting, and
$\rho = 0.421$, obtained by Qwen 2.5 72B for coherence under zero-shot prompting.
These values are higher than the corresponding $\kappa_l$ of 0.069 and 0.171
because $\rho$ reflects only the relative ordering of responses, whereas
$\kappa_l$ also requires the assigned scores to match.
Coherence yields the highest agreement across models, whereas suitableness
remains close to zero, with $\rho$ not exceeding 0.053. Llama 3.1 8B,
Mistral 22B, and both Command R models assign a score of 1 to every instance in
47 of their 48 configurations, which yields $\kappa_l = 0$, so they perform no
better than a majority baseline. Only 38 of the 144 configurations exceed the 95\% upper bound of the random baseline, which is $\kappa_l = 0.023$ for relevance, $0.015$ for coherence, $0.033$ for aggressiveness, and $0.035$ for suitableness. No configuration exceeds this bound for suitableness. While prior work reports rank correlations, such as Spearman's $\rho$ and Kendall's $\tau$, between LLM and human judgments of counterspeech~\cite{jones2024multi,hengle2025cseval}, our results show that moderate rank correlations, with $\rho$ up to 0.43, can coexist with low agreement on the scores themselves, with $\kappa_l$ not exceeding 0.19.

No prompting strategy yields consistently higher agreement in pointwise
evaluation. A Friedman test shows no significant difference in $\kappa_l$
between zero-shot, few-shot, and CoT prompting, $p = 0.231$. Descriptively,
zero-shot prompting yields the highest mean $\kappa_l$ across models and
dimensions, 0.022, followed by few-shot prompting with 0.018 and CoT prompting
with 0.013, and zero-shot prompting gives the highest $\kappa_l$ for three of
the four dimensions, with few-shot prompting doing so for coherence. CoT
prompting yields the lowest agreement, as many models assign the same score to
every instance under this strategy.

Larger models tend to achieve slightly higher agreement than the smaller model of
the same family. A Wilcoxon signed-rank test comparing the two variants of each
family under the same prompting strategy and dimension shows that this
difference is not significant, $p = 0.064$, with the larger model achieving
higher $\kappa_l$ in 32 of 72 cases and lower $\kappa_l$ in 20. Lastly,
comparing the mean $\kappa_l$ of the three size groups shows that agreement does
not increase steadily with size: medium models achieve the highest group mean for
relevance, aggressiveness, and suitableness, and large models only for coherence.
Under CoT prompting, all large models assign the same score to every instance
for relevance, coherence, and aggressiveness, which lowers their mean.

\begin{table}[!htbp]
\centering
\caption{Pointwise human--LLM agreement. Avg.: group mean; $\rho$ is averaged
over models for which it is defined. --: same score for every instance. Bold:
highest $\kappa_l$ per dimension. Majority: most frequent expert score per
dimension. Random (95\% UB): upper end of the range covering 95\% of the
$\kappa_l$ values from 2,000 runs of random scoring; higher values are unlikely
to arise by chance.}
\label{tab:pointwise-full}
\scriptsize\renewcommand{\arraystretch}{0.85}
\setlength{\tabcolsep}{1pt}
\begin{tabular*}{\columnwidth}{@{}l@{\extracolsep{\fill}}lrrrrrrrr@{}}
\toprule
 & & \multicolumn{2}{c}{Relevance} &
\multicolumn{2}{c}{Coherence} &
\multicolumn{2}{c}{Aggressiveness} &
\multicolumn{2}{c}{Suitableness} \\
\cmidrule(lr){3-4}\cmidrule(lr){5-6}\cmidrule(lr){7-8}\cmidrule(lr){9-10}
Model & Prompt & $\kappa_l$ & $\rho$ & $\kappa_l$ & $\rho$ & $\kappa_l$ & $\rho$ & $\kappa_l$ & $\rho$ \\
\midrule
OLMo 7B & Zero-shot & .026 & .065 & .077 & .264 & .001 & .014 & .000 & -- \\
 & Few-shot & .060 & .048 & .058 & .201 & $-$.025 & $-$.067 & $-$.098 & $-$.142 \\
 & CoT & .062 & .125 & .143 & .320 & .000 & -- & .000 & -- \\
\addlinespace[1pt]
Qwen 7B & Zero-shot & .070 & .122 & .161 & .378 & .064 & .307 & $-$.021 & $-$.037 \\
 & Few-shot & .037 & .067 & .149 & .320 & .063 & .296 & $-$.056 & $-$.101 \\
 & CoT & .000 & -- & .000 & -- & .000 & -- & .000 & -- \\
\addlinespace[1pt]
Mistral 7B & Zero-shot & .001 & .028 & $-$.029 & $-$.071 & .003 & .031 & $-$.001 & $-$.026 \\
 & Few-shot & $-$.005 & $-$.017 & $-$.141 & $-$.232 & .000 & $-$.035 & $-$.001 & $-$.026 \\
 & CoT & .000 & -- & .003 & .049 & $-$.001 & $-$.038 & .001 & .020 \\
\addlinespace[1pt]
Cmd R 7B & Zero-shot & .000 & -- & .000 & -- & .000 & -- & .000 & -- \\
 & Few-shot & .000 & -- & .000 & -- & .000 & -- & .000 & -- \\
 & CoT & .000 & -- & .000 & -- & .000 & -- & .000 & -- \\
\addlinespace[1pt]
Llama 8B & Zero-shot & .000 & -- & .000 & -- & .000 & -- & .000 & -- \\
 & Few-shot & .000 & -- & .000 & -- & .000 & -- & .000 & -- \\
 & CoT & .000 & -- & .000 & -- & .000 & -- & .000 & -- \\
\addlinespace[1pt]
\textit{Small avg.} & Zero-shot & .019 & .072 & .042 & .190 & .014 & .117 & $-$.004 & $-$.032 \\
 & Few-shot & .018 & .033 & .013 & .096 & .008 & .065 & $-$.031 & $-$.090 \\
 & CoT & .012 & .125 & .029 & .184 & .000 & $-$.038 & .000 & .020 \\
\midrule
Mixtral 8$\times$7B & Zero-shot & .004 & .219 & .027 & .404 & \textbf{.106} & .375 & .029 & $-$.014 \\
 & Few-shot & .058 & .100 & .059 & .313 & .016 & .072 & $-$.002 & $-$.002 \\
 & CoT & .002 & .055 & .130 & .355 & .074 & .258 & $-$.024 & $-$.066 \\
\addlinespace[1pt]
Mistral 22B & Zero-shot & .000 & -- & .000 & -- & .000 & -- & .000 & -- \\
 & Few-shot & .000 & -- & .000 & -- & .000 & -- & .000 & -- \\
 & CoT & .000 & -- & .000 & -- & .000 & -- & .000 & -- \\
\addlinespace[1pt]
OLMo 32B & Zero-shot & .024 & .198 & .136 & .368 & .042 & .273 & \textbf{.031} & .053 \\
 & Few-shot & .052 & .132 & .103 & .371 & .095 & .258 & $-$.015 & $-$.024 \\
 & CoT & .005 & .028 & .188 & .382 & .009 & .178 & .000 & -- \\
\addlinespace[1pt]
Mixtral 8$\times$22B & Zero-shot & .023 & .204 & .037 & .337 & .069 & .431 & $-$.008 & $-$.122 \\
 & Few-shot & .043 & .194 & .042 & .338 & .016 & .164 & .011 & $-$.013 \\
 & CoT & .000 & -- & .003 & .028 & .014 & .194 & .000 & -- \\
\addlinespace[1pt]
\textit{Medium avg.} & Zero-shot & .013 & .207 & .050 & .370 & .054 & .360 & .013 & $-$.028 \\
 & Few-shot & .038 & .142 & .051 & .341 & .032 & .165 & $-$.002 & $-$.013 \\
 & CoT & .002 & .042 & .080 & .255 & .024 & .210 & $-$.006 & $-$.066 \\
\midrule
Llama 70B & Zero-shot & $-$.022 & $-$.132 & .002 & .029 & .000 & .006 & .001 & .008 \\
 & Few-shot & .047 & .057 & .099 & .212 & .009 & .163 & $-$.071 & $-$.093 \\
 & CoT & .000 & -- & .000 & -- & .000 & -- & .005 & .053 \\
\addlinespace[1pt]
Qwen 72B & Zero-shot & \textbf{.096} & .151 & .171 & .421 & .105 & .387 & $-$.191 & $-$.266 \\
 & Few-shot & .046 & .027 & \textbf{.191} & .359 & .087 & .187 & $-$.046 & $-$.071 \\
 & CoT & .000 & -- & .000 & -- & .000 & -- & .000 & -- \\
\addlinespace[1pt]
Cmd R+ 104B & Zero-shot & .000 & -- & .000 & -- & .000 & -- & .000 & -- \\
 & Few-shot & $-$.004 & $-$.071 & .000 & -- & .000 & -- & .000 & -- \\
 & CoT & .000 & -- & .000 & -- & .000 & -- & .000 & -- \\
\addlinespace[1pt]
\textit{Large avg.} & Zero-shot & .025 & .009 & .058 & .225 & .035 & .197 & $-$.063 & $-$.129 \\
 & Few-shot & .030 & .004 & .097 & .285 & .032 & .175 & $-$.039 & $-$.082 \\
 & CoT & .000 & -- & .000 & -- & .000 & -- & .002 & .053 \\
\midrule
\textit{Majority} & -- & .000 & -- & .000 & -- & .000 & -- & .000 & -- \\
\textit{Random (95\% UB)} & -- & .023 & .056 & .015 & .052 & .033 & .059 & .035 & .054 \\
\bottomrule
\end{tabular*}
\end{table}

\subsection{Human--LLM Agreement in Pairwise Evaluation}
\label{sec:results-pairwise}

Tables~\ref{tab:pairwise-same} and~\ref{tab:pairwise-cross} report agreement
for the same 144 configurations in same- and cross-instance comparisons,
respectively. 

\begin{table}[!htbp]
\centering
\caption{Same-instance comparisons: pairwise human--LLM agreement. Bold: highest
$\kappa$ and $\kappa_{\mathrm{pw}}$ per dimension. Avg.: group mean. Majority and
Random (95\% UB): as in Table~\ref{tab:pointwise-full}.}
\label{tab:pairwise-same}
\scriptsize\renewcommand{\arraystretch}{0.85}
\setlength{\tabcolsep}{1pt}
\begin{tabular*}{\columnwidth}{@{}l@{\extracolsep{\fill}}lrrrrrrrr@{}}
\toprule
 & & \multicolumn{2}{c}{Relevance} &
\multicolumn{2}{c}{Coherence} &
\multicolumn{2}{c}{Aggressiveness} &
\multicolumn{2}{c}{Suitableness} \\
\cmidrule(lr){3-4}\cmidrule(lr){5-6}\cmidrule(lr){7-8}\cmidrule(lr){9-10}
Model & Prompt & $\kappa$ & $\kappa_{\mathrm{pw}}$ & $\kappa$ & $\kappa_{\mathrm{pw}}$ & $\kappa$ & $\kappa_{\mathrm{pw}}$ & $\kappa$ & $\kappa_{\mathrm{pw}}$ \\
\midrule
OLMo 7B & Zero-shot & .080 & .040 & .159 & .216 & .010 & $-$.003 & $-$.014 & .000 \\
 & Few-shot & .008 & .150 & $-$.003 & .217 & $-$.002 & .069 & $-$.001 & $-$.049 \\
 & CoT & .091 & .086 & .099 & .264 & .056 & .000 & $-$.001 & .000 \\
\addlinespace[1pt]
Qwen 7B & Zero-shot & .073 & .017 & .192 & .305 & .010 & .176 & $-$.019 & .036 \\
 & Few-shot & .056 & .097 & .178 & .251 & .029 & .169 & $-$.030 & .036 \\
 & CoT & .083 & .000 & .195 & .000 & .077 & .000 & $-$.016 & .000 \\
\addlinespace[1pt]
Mistral 7B & Zero-shot & .111 & .000 & .167 & $-$.025 & $-$.008 & $-$.006 & \textbf{.035} & .002 \\
 & Few-shot & .039 & $-$.016 & .055 & $-$.063 & .033 & $-$.048 & .002 & .000 \\
 & CoT & .040 & .000 & .163 & .000 & .072 & .000 & .014 & .000 \\
\addlinespace[1pt]
Cmd R 7B & Zero-shot & .064 & .000 & .185 & .000 & .036 & .000 & .017 & .000 \\
 & Few-shot & .031 & .000 & .181 & .000 & .000 & .000 & $-$.020 & .000 \\
 & CoT & .111 & .000 & .218 & .000 & .054 & .000 & .001 & .000 \\
\addlinespace[1pt]
Llama 8B & Zero-shot & .008 & .000 & .078 & .000 & .000 & .000 & $-$.008 & .000 \\
 & Few-shot & $-$.005 & .000 & .075 & .000 & .040 & .000 & .024 & .000 \\
 & CoT & .086 & .000 & .196 & .000 & .014 & .000 & $-$.006 & .000 \\
\addlinespace[1pt]
\textit{Small avg.} & Zero-shot & .067 & .011 & .156 & .099 & .010 & .033 & .002 & .008 \\
 & Few-shot & .026 & .046 & .097 & .081 & .020 & .038 & $-$.005 & $-$.003 \\
 & CoT & .082 & .017 & .174 & .053 & .055 & .000 & $-$.002 & .000 \\
\midrule
Mixtral 8$\times$7B & Zero-shot & .082 & .104 & .207 & .305 & $-$.002 & .163 & .013 & .032 \\
 & Few-shot & $-$.004 & .091 & .074 & .343 & .009 & .025 & $-$.045 & .000 \\
 & CoT & .092 & .068 & .206 & .233 & .066 & .051 & .009 & $-$.031 \\
\addlinespace[1pt]
Mistral 22B & Zero-shot & .033 & .000 & .017 & .000 & .000 & .000 & .000 & .000 \\
 & Few-shot & .049 & .000 & .043 & .000 & $-$.001 & .000 & .001 & .000 \\
 & CoT & .003 & .000 & .012 & .000 & .000 & .000 & .009 & .000 \\
\addlinespace[1pt]
OLMo 32B & Zero-shot & .088 & .111 & .214 & .284 & \textbf{.159} & .134 & $-$.004 & $-$.012 \\
 & Few-shot & .093 & .118 & .203 & .305 & .118 & .223 & $-$.012 & $-$.020 \\
 & CoT & .100 & $-$.021 & .204 & .276 & .130 & .019 & $-$.007 & .000 \\
\addlinespace[1pt]
Mixtral 8$\times$22B & Zero-shot & .100 & \textbf{.215} & .226 & .245 & .116 & .197 & .005 & .014 \\
 & Few-shot & \textbf{.174} & .150 & .174 & .275 & .019 & .056 & $-$.011 & \textbf{.071} \\
 & CoT & .098 & .000 & .177 & .029 & .132 & .068 & $-$.018 & .000 \\
\addlinespace[1pt]
\textit{Medium avg.} & Zero-shot & .076 & .107 & .166 & .208 & .068 & .123 & .003 & .009 \\
 & Few-shot & .078 & .090 & .123 & .231 & .036 & .076 & $-$.017 & .013 \\
 & CoT & .073 & .012 & .150 & .135 & .082 & .035 & $-$.002 & $-$.008 \\
\midrule
Llama 70B & Zero-shot & .101 & $-$.032 & .181 & .000 & .098 & .000 & $-$.018 & $-$.010 \\
 & Few-shot & .086 & .149 & .170 & .257 & .061 & .052 & $-$.016 & $-$.012 \\
 & CoT & .095 & .000 & .179 & .000 & .073 & .000 & $-$.021 & .019 \\
\addlinespace[1pt]
Qwen 72B & Zero-shot & .104 & .185 & \textbf{.245} & \textbf{.353} & .061 & \textbf{.234} & .001 & $-$.078 \\
 & Few-shot & .089 & .128 & .214 & .292 & .082 & .112 & $-$.014 & .015 \\
 & CoT & .108 & .000 & .232 & .000 & .101 & .000 & $-$.011 & .000 \\
\addlinespace[1pt]
Cmd R+ 104B & Zero-shot & .102 & .000 & .205 & .000 & $-$.034 & .000 & .015 & .000 \\
 & Few-shot & .134 & .003 & .196 & .000 & $-$.013 & .000 & .015 & .000 \\
 & CoT & .102 & .000 & .188 & .000 & $-$.025 & .000 & $-$.012 & .000 \\
\addlinespace[1pt]
\textit{Large avg.} & Zero-shot & .102 & .051 & .210 & .118 & .042 & .078 & $-$.001 & $-$.029 \\
 & Few-shot & .103 & .093 & .193 & .183 & .043 & .055 & $-$.005 & .001 \\
 & CoT & .102 & .000 & .200 & .000 & .050 & .000 & $-$.015 & .006 \\
\midrule
\textit{Majority} & -- & .000 & .000 & .000 & .000 & .000 & .000 & .000 & .000 \\
\textit{Random (95\% UB)} & -- & .062 & .062 & .065 & .065 & .082 & .082 & .063 & .063 \\
\bottomrule
\end{tabular*}
\end{table}

\begin{table}[!htbp]
\centering
\caption{Cross-instance comparisons: pairwise human--LLM agreement. Bold: highest
$\kappa$ and $\kappa_{\mathrm{pw}}$ per dimension. Avg.: group mean. Majority and
Random (95\% UB): as in Table~\ref{tab:pointwise-full}.}
\label{tab:pairwise-cross}
\scriptsize\renewcommand{\arraystretch}{0.85}
\setlength{\tabcolsep}{1pt}
\begin{tabular*}{\columnwidth}{@{}l@{\extracolsep{\fill}}lrrrrrrrr@{}}
\toprule
 & & \multicolumn{2}{c}{Relevance} &
\multicolumn{2}{c}{Coherence} &
\multicolumn{2}{c}{Aggressiveness} &
\multicolumn{2}{c}{Suitableness} \\
\cmidrule(lr){3-4}\cmidrule(lr){5-6}\cmidrule(lr){7-8}\cmidrule(lr){9-10}
Model & Prompt & $\kappa$ & $\kappa_{\mathrm{pw}}$ & $\kappa$ & $\kappa_{\mathrm{pw}}$ & $\kappa$ & $\kappa_{\mathrm{pw}}$ & $\kappa$ & $\kappa_{\mathrm{pw}}$ \\
\midrule
OLMo 7B & Zero-shot & .029 & .012 & $-$.025 & .037 & .011 & $-$.003 & $-$.011 & .000 \\
 & Few-shot & $-$.001 & $-$.012 & .000 & $-$.046 & .002 & $-$.008 & .000 & .002 \\
 & CoT & $-$.004 & .045 & $-$.006 & .029 & .028 & .000 & $-$.005 & .000 \\
\addlinespace[1pt]
Qwen 7B & Zero-shot & .038 & .070 & .042 & .037 & .039 & .129 & .011 & .025 \\
 & Few-shot & .033 & .059 & .036 & .034 & $-$.016 & .119 & $-$.023 & .054 \\
 & CoT & .024 & .000 & .040 & .000 & .061 & .000 & .008 & .000 \\
\addlinespace[1pt]
Mistral 7B & Zero-shot & .002 & $-$.002 & .046 & $-$.066 & $-$.014 & $-$.002 & $-$.014 & .000 \\
 & Few-shot & .008 & $-$.007 & .031 & $-$.034 & $-$.001 & $-$.037 & $-$.010 & .000 \\
 & CoT & $-$.004 & .000 & .043 & .000 & $-$.004 & .000 & $-$.006 & .000 \\
\addlinespace[1pt]
Cmd R 7B & Zero-shot & .012 & .000 & .006 & .000 & .002 & .000 & .004 & .000 \\
 & Few-shot & .032 & .000 & .024 & .000 & .017 & .000 & .006 & .000 \\
 & CoT & .003 & .000 & .001 & .000 & $-$.004 & .000 & .000 & .000 \\
\addlinespace[1pt]
Llama 8B & Zero-shot & .001 & .000 & .003 & .000 & .000 & .000 & .000 & .000 \\
 & Few-shot & $-$.001 & .000 & .001 & .000 & $-$.002 & .000 & $-$.002 & .000 \\
 & CoT & .000 & .000 & .000 & .000 & .000 & .000 & .000 & .000 \\
\addlinespace[1pt]
\textit{Small avg.} & Zero-shot & .016 & .016 & .014 & .002 & .008 & .025 & $-$.002 & .005 \\
 & Few-shot & .014 & .008 & .018 & $-$.009 & .000 & .015 & $-$.006 & .011 \\
 & CoT & .004 & .009 & .016 & .006 & .016 & .000 & $-$.001 & .000 \\
\midrule
Mixtral 8$\times$7B & Zero-shot & .049 & .054 & .039 & .066 & .010 & .149 & $-$.014 & .038 \\
 & Few-shot & .002 & .058 & .005 & .024 & .000 & .045 & $-$.048 & .014 \\
 & CoT & .041 & .020 & .045 & .044 & .020 & .077 & .009 & $-$.011 \\
\addlinespace[1pt]
Mistral 22B & Zero-shot & .010 & .000 & $-$.005 & .000 & .002 & .000 & $-$.008 & .000 \\
 & Few-shot & .014 & .000 & $-$.022 & .000 & .018 & .000 & $-$.010 & .000 \\
 & CoT & .003 & .000 & $-$.004 & .000 & .003 & .000 & $-$.004 & .000 \\
\addlinespace[1pt]
OLMo 32B & Zero-shot & .050 & .045 & .034 & .061 & .078 & .076 & .031 & .071 \\
 & Few-shot & .041 & .037 & .055 & \textbf{.088} & .071 & .143 & \textbf{.042} & .071 \\
 & CoT & .049 & .005 & .032 & .060 & .091 & .027 & .037 & .000 \\
\addlinespace[1pt]
Mixtral 8$\times$22B & Zero-shot & \textbf{.107} & \textbf{.082} & .051 & .018 & .091 & .143 & .041 & .019 \\
 & Few-shot & .066 & .078 & .046 & .007 & .063 & .023 & .028 & \textbf{.126} \\
 & CoT & .092 & .000 & .037 & .001 & .107 & .016 & .039 & .000 \\
\addlinespace[1pt]
\textit{Medium avg.} & Zero-shot & .054 & .045 & .030 & .036 & .045 & .092 & .013 & .032 \\
 & Few-shot & .031 & .043 & .021 & .030 & .038 & .053 & .003 & .053 \\
 & CoT & .046 & .006 & .028 & .026 & .055 & .030 & .020 & $-$.003 \\
\midrule
Llama 70B & Zero-shot & .046 & $-$.017 & .010 & .005 & .111 & $-$.001 & $-$.009 & .063 \\
 & Few-shot & .035 & .049 & .018 & $-$.001 & .100 & .030 & $-$.002 & .074 \\
 & CoT & .037 & .000 & .025 & .000 & .093 & .000 & $-$.005 & .001 \\
\addlinespace[1pt]
Qwen 72B & Zero-shot & .067 & .075 & \textbf{.059} & .076 & .102 & \textbf{.151} & .025 & .013 \\
 & Few-shot & .022 & .079 & .041 & .036 & \textbf{.130} & .113 & .012 & .003 \\
 & CoT & .063 & .000 & .051 & .000 & .128 & .000 & $-$.003 & .000 \\
\addlinespace[1pt]
Cmd R+ 104B & Zero-shot & .003 & .000 & .007 & .000 & $-$.001 & .000 & .005 & .000 \\
 & Few-shot & .030 & $-$.001 & .041 & .000 & $-$.021 & .000 & .011 & .000 \\
 & CoT & .004 & .000 & .009 & .000 & .005 & .000 & .007 & .000 \\
\addlinespace[1pt]
\textit{Large avg.} & Zero-shot & .039 & .019 & .025 & .027 & .071 & .050 & .007 & .025 \\
 & Few-shot & .029 & .042 & .033 & .012 & .070 & .048 & .007 & .026 \\
 & CoT & .035 & .000 & .028 & .000 & .075 & .000 & .000 & .000 \\
\midrule
\textit{Majority} & -- & .000 & .000 & .000 & .000 & .000 & .000 & .000 & .000 \\
\textit{Random (95\% UB)} & -- & .016 & .016 & .016 & .016 & .017 & .017 & .017 & .017 \\
\bottomrule
\end{tabular*}
\end{table}

No pairwise configuration exceeds fair agreement in same-instance
comparisons or slight agreement in cross-instance comparisons on the
Landis--Koch scale. In same-instance comparisons, the highest
agreement per dimension is obtained for coherence by Qwen 2.5 72B under
zero-shot, with $\kappa = 0.245$, 95\% CI [0.198, 0.301], for relevance by
Mixtral 8$\times$22B under few-shot, with $\kappa = 0.174$, for aggressiveness
by OLMo 2 32B under zero-shot, with $\kappa = 0.159$, and for suitableness by
Mistral 7B under zero-shot, with $\kappa = 0.035$. Coherence is the only
dimension that reaches fair agreement, for Qwen 2.5 72B under all three
prompting strategies and for Mixtral 8$\times$22B, OLMo 2 32B, and Command R 7B
under one strategy each. In cross-instance comparisons, the highest agreement
per dimension is obtained for aggressiveness by Qwen 2.5 72B under few-shot,
with $\kappa = 0.130$ [0.073, 0.186], for relevance by Mixtral 8$\times$22B
under zero-shot, with $\kappa = 0.107$, for coherence by Qwen 2.5 72B under
zero-shot, with $\kappa = 0.059$, and for suitableness by OLMo 2 32B under
few-shot, with $\kappa = 0.042$. As in pointwise evaluation, suitableness yields
the lowest agreement in both comparison types, whereas coherence yields the
highest agreement only in same-instance comparisons. Only 63 of the 144 same-instance configurations exceed the 95\% upper bound of the random baseline, which
is $\kappa = 0.062$ for relevance, $0.065$ for coherence, $0.082$ for
aggressiveness, and $0.063$ for suitableness; no configuration exceeds this bound
for suitableness. Only 65 of the 144 cross-instance configurations exceed this
bound, which is $\kappa = 0.016$ for relevance, $0.016$ for coherence, $0.017$ for
aggressiveness, and $0.017$ for suitableness; only seven configurations exceed
it for suitableness.


No prompting strategy yields consistently higher agreement in pairwise
evaluation either. In same-instance comparisons, few-shot prompting yields lower agreement than the other two strategies: mean $\kappa$ is 0.054 under few-shot, compared with 0.073 under zero-shot and 0.079 under CoT. This difference is significant (Friedman test, $p < 0.001$): zero-shot outperforms few-shot ($p = 0.012$), whereas zero-shot and CoT do not differ ($p = 0.582$), both with the Wilcoxon signed-rank test. In cross-instance comparisons, the three strategies do not differ significantly (Friedman test, $p = 0.408$), with mean $\kappa$ of 0.024
under zero-shot, 0.019 under few-shot, and 0.025 under CoT. The strategy that
yields the highest $\kappa$ also varies by dimension, as reported above: in
cross-instance comparisons, zero-shot for relevance and coherence and few-shot
for aggressiveness and suitableness.

Larger models achieve higher agreement than the smaller model of the same family
in pairwise evaluation. A Wilcoxon signed-rank test comparing the two variants of
each family under the same prompting strategy and dimension shows that this
difference is significant in both same-instance comparisons, $p = 0.003$, with
the larger model achieving higher $\kappa$ in 46 of 72 cases, and cross-instance
comparisons, $p < 0.001$, with the larger model achieving higher $\kappa$ in 59
of 72 cases. Mistral and Command R are the exceptions in same-instance
comparisons, where the smaller model achieves higher $\kappa$ in most cases.
Lastly, large models achieve the highest mean $\kappa$ for relevance and
coherence in same-instance comparisons, and for coherence and aggressiveness in
cross-instance comparisons, while small models achieve the lowest mean $\kappa$
in nearly all cases.

\subsection{Pointwise vs.\ Pairwise Evaluation}
\label{sec:results-setting-comparison}

The preceding results use linear-weighted $\kappa_l$ for pointwise evaluation
and unweighted $\kappa$ for pairwise evaluation. To compare the two settings
directly, we use $\kappa_{\mathrm{pw}}$ in Tables~\ref{tab:pairwise-same}
and~\ref{tab:pairwise-cross}, the agreement of labels converted from each
model's pointwise scores into pairwise labels for the same response pairs
(Section~\ref{sec:methodology}). Since $\kappa_{\mathrm{pw}}$ and $\kappa$ are
both unweighted $\kappa$ against the same reference labels on the same pairs,
they are directly comparable.

A Wilcoxon signed-rank test shows no significant difference between pairwise
and pointwise evaluation, $p = 0.073$ for same-instance and $p = 0.188$ for
cross-instance comparisons. Neither setting is consistently better. This is partly
because many models give the same pointwise score to every instance, which turns
all their pairs into ties and yields $\kappa_{\mathrm{pw}} = 0$. When models do
vary their scores, pointwise evaluation can perform clearly better: for
same-instance coherence under zero-shot prompting, Qwen 2.5 72B reaches
$\kappa_{\mathrm{pw}} = 0.353$, 95\% CI [0.284, 0.425], compared with
$\kappa = 0.245$ for its pairwise judgments, and a paired cluster bootstrap
shows that this difference of 0.107 is significant, 95\% CI [0.053, 0.163].

\subsection{Agreement Between LLMs}
\label{sec:results-llm-llm}

We now examine LLM--LLM agreement, the agreement between two LLMs judging the
same responses. We take OLMo 2 32B as the reference model, since it achieves the
highest mean agreement with the experts across prompting strategies and
dimensions in pointwise evaluation, $\kappa_l = 0.056$, and same-instance
comparisons, $\kappa = 0.107$, and the third highest in cross-instance
comparisons, $\kappa = 0.051$. For each of the three prompting strategies and
each quality dimension, we compare the agreement of OLMo 2 32B with each of the
other 11 LLMs to its agreement with the experts under the same strategy and
dimension. Table~\ref{tab:llm-llm-olmo232b} reports the results per model and
the mean agreement per size group.

LLMs agree more with one another than with the experts in pairwise evaluation.
In same-instance comparisons, OLMo 2 32B reaches a mean $\kappa = 0.390$ with
the other LLMs, compared with $\kappa = 0.107$ with the experts, and agrees more
with another LLM than with the experts in 109 of 132 comparisons. In
cross-instance comparisons, it reaches a mean $\kappa = 0.207$ with the other
LLMs, compared with $\kappa = 0.051$ with the experts, and agrees more with
another LLM in 94 of 132 comparisons. In pointwise evaluation,
the difference is smaller: its mean agreement with the other LLMs is
$\kappa_l = 0.118$, and it agrees more with another LLM in only 50 of 124
comparisons, partly because many models assign the same pointwise score to every
instance, which yields zero agreement. Pairwise agreement tends to increase with
the size of the other model: for same-instance coherence under zero-shot
prompting, OLMo 2 32B reaches a mean $\kappa = 0.717$ with large models,
compared with $0.477$ with medium and $0.440$ with small models, and its highest
agreement, $\kappa = 0.764$, 95\% CI [0.710, 0.827], is with Llama 3.3 70B,
whereas it reaches only $\kappa = 0.214$ with the experts in the same
configuration. The gap is largest for suitableness: agreement with the experts
on this dimension does not exceed $\kappa = 0.042$ in any setting, whereas
agreement with other LLMs reaches $\kappa_l = 0.633$ with Qwen 2.5 72B under
few-shot in pointwise evaluation, $\kappa = 0.746$ with Qwen 2.5 72B
under CoT in same-instance comparisons, and $\kappa = 0.537$ with
Mixtral 8$\times$22B under zero-shot in cross-instance comparisons.

\begin{table}[H]
\centering
\caption{Agreement of OLMo 2 32B with the experts and each of the other 11 LLMs.
Avg.: group mean, excluding OLMo 2 32B. --: both models give the same score to
every instance. Bold: highest LLM--LLM $\kappa$ per column.
Rel./Coh./Agg./Suit.: relevance, coherence, aggressiveness, suitableness.}
\label{tab:llm-llm-olmo232b}
\scriptsize\renewcommand{\arraystretch}{0.85}
\setlength{\tabcolsep}{1.2pt}
\begin{adjustbox}{max width=\columnwidth}
\begin{tabular}{ll*{4}{r}@{\hspace{4pt}}*{4}{r}@{\hspace{4pt}}*{4}{r}}
\toprule
 & & \multicolumn{4}{c}{Pointwise} & \multicolumn{4}{c}{Same-instance} & \multicolumn{4}{c}{Cross-instance} \\
\cmidrule(lr){3-6}\cmidrule(lr){7-10}\cmidrule(lr){11-14}
Model & Prompt & Rel. & Coh. & Agg. & Suit. &  Rel. & Coh. & Agg. & Suit. &  Rel. & Coh. & Agg. & Suit. \\
\midrule
\textit{Experts} & Zero-shot & .024 & .136 & .042 & .031 & .088 & .214 & .159 & $-$.004 & .050 & .034 & .078 & .031 \\
 & Few-shot & .052 & .103 & .095 & $-$.015 & .093 & .203 & .118 & $-$.012 & .041 & .055 & .071 & .042 \\
 & CoT & .005 & .188 & .009 & .000 & .100 & .204 & .130 & $-$.007 & .049 & .032 & .091 & .037 \\
\midrule
OLMo 7B & Zero-shot & .136 & .449 & .006 & .000 & .374 & .461 & .020 & .427 & .333 & .295 & .068 & .169 \\
 & Few-shot & .322 & .481 & .019 & .231 & .016 & .009 & .008 & .003 & .001 & .002 & .000 & .000 \\
 & CoT & .073 & \textbf{.652} & .000 & -- & .181 & .247 & .050 & .208 & .208 & .034 & .028 & .038 \\
\addlinespace[1pt]
Qwen 7B & Zero-shot & .016 & .570 & .351 & .472 & .625 & .657 & .136 & .663 & .432 & \textbf{.501} & .227 & .405 \\
 & Few-shot & .091 & .393 & .162 & .471 & .568 & .616 & .234 & .533 & .344 & .233 & .157 & .338 \\
 & CoT & .000 & .000 & .000 & -- & .547 & .681 & .320 & .656 & .283 & .364 & .317 & .426 \\
\addlinespace[1pt]
Mistral 7B & Zero-shot & .000 & .003 & .194 & .001 & .346 & .391 & .172 & .217 & .195 & .307 & .198 & .151 \\
 & Few-shot & .001 & $-$.097 & $-$.077 & .001 & .501 & .199 & .208 & .394 & .100 & .298 & .083 & .110 \\
 & CoT & .000 & .003 & .128 & .000 & .217 & .484 & .327 & .294 & .067 & .202 & .187 & .138 \\
\addlinespace[1pt]
Cmd R 7B & Zero-shot & .000 & .000 & .000 & .000 & .493 & .551 & .109 & .579 & .008 & .006 & .003 & .009 \\
 & Few-shot & .000 & .000 & .000 & .000 & .311 & .456 & .099 & .350 & .051 & .033 & .046 & .043 \\
 & CoT & .000 & .000 & .000 & -- & .325 & .494 & .116 & .447 & .002 & .010 & .002 & .004 \\
\addlinespace[1pt]
Llama 8B & Zero-shot & .000 & .000 & .000 & .000 & .014 & .139 & .011 & .039 & .006 & .008 & .000 & .002 \\
 & Few-shot & .000 & .000 & .000 & .000 & .095 & .192 & .291 & .217 & .052 & .009 & .012 & .008 \\
 & CoT & .000 & .000 & .000 & -- & .163 & .333 & .021 & .089 & .002 & $-$.001 & $-$.001 & .000 \\
\addlinespace[1pt]
\textit{Small avg.} & Zero-shot & .030 & .204 & .110 & .095 & .370 & .440 & .090 & .385 & .195 & .223 & .099 & .147 \\
 & Few-shot & .083 & .155 & .021 & .141 & .298 & .294 & .168 & .299 & .110 & .115 & .060 & .100 \\
 & CoT & .015 & .131 & .026 & .000 & .287 & .448 & .167 & .339 & .112 & .122 & .107 & .121 \\
\midrule
Mixtral 8$\times$7B & Zero-shot & .152 & .289 & \textbf{.436} & .358 & .674 & .687 & .202 & .695 & .371 & .372 & .180 & .163 \\
 & Few-shot & .280 & .446 & .114 & .003 & .384 & .234 & .294 & .455 & .125 & .073 & .022 & .105 \\
 & CoT & .002 & .468 & .190 & .000 & .561 & .633 & .344 & .692 & .255 & .226 & .174 & .212 \\
\addlinespace[1pt]
Mistral 22B & Zero-shot & .000 & .000 & .000 & .000 & .195 & .038 & .000 & .163 & .065 & .021 & .001 & .065 \\
 & Few-shot & .000 & .000 & .000 & .000 & .210 & .124 & .009 & .215 & .123 & .027 & .071 & .080 \\
 & CoT & .000 & .000 & .000 & -- & .011 & .027 & .000 & .042 & .013 & .008 & .003 & .017 \\
\addlinespace[1pt]
Mixtral 8$\times$22B & Zero-shot & .467 & .376 & .275 & .345 & .654 & .706 & \textbf{.565} & .719 & \textbf{.505} & .482 & .437 & \textbf{.537} \\
 & Few-shot & \textbf{.559} & .484 & .059 & .594 & .501 & .462 & .293 & .509 & .332 & .371 & .261 & .358 \\
 & CoT & .000 & .003 & .348 & -- & .552 & .590 & \textbf{.565} & .648 & .462 & .467 & .450 & .499 \\
\addlinespace[1pt]
\textit{Medium avg.} & Zero-shot & .206 & .222 & .237 & .234 & .508 & .477 & .256 & .526 & .314 & .292 & .206 & .255 \\
 & Few-shot & .280 & .310 & .058 & .199 & .365 & .273 & .199 & .393 & .193 & .157 & .118 & .181 \\
 & CoT & .001 & .157 & .179 & .000 & .375 & .417 & .303 & .461 & .243 & .234 & .209 & .243 \\
\midrule
Llama 70B & Zero-shot & $-$.009 & $-$.001 & $-$.002 & $-$.060 & .736 & \textbf{.764} & .357 & .742 & .461 & .473 & .431 & .438 \\
 & Few-shot & .108 & .218 & .048 & .158 & .622 & .624 & .274 & .522 & .333 & .411 & .295 & .470 \\
 & CoT & .000 & .000 & .000 & .000 & .564 & .670 & .246 & .667 & .290 & .406 & .386 & .394 \\
\addlinespace[1pt]
Qwen 72B & Zero-shot & .177 & .493 & .370 & .280 & \textbf{.746} & .759 & .367 & .745 & .481 & .422 & .406 & .489 \\
 & Few-shot & .223 & .346 & .296 & \textbf{.633} & .613 & .647 & .453 & .658 & .332 & .360 & .332 & .372 \\
 & CoT & .000 & .000 & .000 & -- & .607 & .741 & .448 & \textbf{.746} & .386 & .437 & \textbf{.494} & .431 \\
\addlinespace[1pt]
Cmd R+ 104B & Zero-shot & .000 & .000 & .000 & .000 & .513 & .629 & .126 & .563 & .126 & .304 & $-$.001 & .120 \\
 & Few-shot & .000 & .000 & .000 & .000 & .527 & .598 & .229 & .521 & .332 & .319 & .162 & .385 \\
 & CoT & .000 & .000 & .000 & -- & .507 & .611 & .178 & .600 & .060 & .130 & .001 & .121 \\
\addlinespace[1pt]
\textit{Large avg.} & Zero-shot & .056 & .164 & .123 & .073 & .665 & .717 & .283 & .683 & .356 & .400 & .279 & .349 \\
 & Few-shot & .110 & .188 & .115 & .264 & .587 & .623 & .319 & .567 & .332 & .363 & .263 & .409 \\
 & CoT & .000 & .000 & .000 & .000 & .559 & .674 & .291 & .671 & .245 & .324 & .294 & .315 \\
\bottomrule
\end{tabular}
\end{adjustbox}
\end{table}

\subsection{Human--LLM Agreement Across Target Groups}
\label{sec:target-groups}

We examine whether agreement differs across the groups targeted by the hate
speech, using OLMo 2 32B under zero-shot prompting, with target groups
as described in Section~\ref{sec:methodology-data}.
Table~\ref{tab:target-groups} reports agreement per target group for all
dimensions and settings, with the number of annotated instances or pairs per
dimension.

\begin{table}[!htbp]
\centering
\caption{Human--LLM agreement of OLMo 2 32B per target group under zero-shot
prompting. $n$: instances or pairs annotated for the dimension. Bold: highest
$\kappa_l$ or $\kappa$ per dimension and setting.}
\label{tab:target-groups}
\renewcommand{\arraystretch}{0.85}
\setlength{\tabcolsep}{1.5pt}
\resizebox{\columnwidth}{!}{%
\begin{tabular}{@{}lrrrrrrrrrrrr@{}}
\toprule
 & \multicolumn{3}{c}{Relevance} & \multicolumn{3}{c}{Coherence} &
\multicolumn{3}{c}{Aggressiveness} & \multicolumn{3}{c}{Suitableness} \\
\cmidrule(lr){2-4}\cmidrule(lr){5-7}\cmidrule(lr){8-10}\cmidrule(lr){11-13}
\textit{Pointwise} & $n$ & $\kappa_l$ & $\rho$ & $n$ & $\kappa_l$ & $\rho$ & $n$ & $\kappa_l$ & $\rho$ & $n$ & $\kappa_l$ & $\rho$ \\
\addlinespace[1pt]
All & 1389 & .024 & .198 & 1397 & .136 & .368 & 1092 & .042 & .273 & 1386 & .031 & .053 \\
\addlinespace[1pt]
Muslims & 460 & .013 & .148 & 463 & .132 & .363 & 393 & .012 & .099 & 459 & .057 & .112 \\
Women & 224 & .023 & .207 & 226 & .083 & .250 & 168 & .069 & .418 & 225 & $-$.025 & $-$.049 \\
Black people & 173 & .030 & .126 & 176 & .149 & .491 & 89 & .069 & .264 & 175 & $-$.024 & $-$.042 \\
LGBTQ+ & 151 & \textbf{.056} & .354 & 151 & .165 & .433 & 113 & .065 & .344 & 149 & .049 & .073 \\
Immigrants & 135 & .016 & .114 & 135 & .091 & .226 & 117 & .049 & .307 & 134 & .007 & .019 \\
Jews & 117 & .015 & .144 & 117 & .164 & .379 & 114 & .019 & .237 & 118 & \textbf{.098} & .162 \\
Other & 129 & .038 & .367 & 129 & \textbf{.178} & .451 & 98 & \textbf{.113} & .376 & 126 & .041 & .090 \\
\midrule
\textit{Same-inst.} & $n$ & $\kappa$ & $\kappa_{\mathrm{pw}}$ & $n$ & $\kappa$ & $\kappa_{\mathrm{pw}}$ & $n$ & $\kappa$ & $\kappa_{\mathrm{pw}}$ & $n$ & $\kappa$ & $\kappa_{\mathrm{pw}}$ \\
\addlinespace[1pt]
All & 441 & .088 & .111 & 450 & .214 & .284 & 263 & .159 & .134 & 447 & $-$.004 & $-$.012 \\
\addlinespace[1pt]
Muslims & 174 & .029 & .090 & 176 & .194 & .248 & 129 & .109 & .109 & 172 & $-$.006 & $-$.002 \\
Women & 69 & .046 & .313 & 72 & .136 & .163 & 34 & .310 & .038 & 72 & $-$.078 & $-$.124 \\
Black people & 68 & .039 & $-$.051 & 71 & \textbf{.325} & .473 & 21 & .125 & $-$.024 & 72 & \textbf{.059} & .062 \\
LGBTQ+ & 55 & \textbf{.406} & .156 & 55 & .289 & .363 & 27 & .211 & .285 & 55 & .004 & .020 \\
Immigrants & 16 & .111 & .075 & 16 & .264 & .264 & 10 & \textbf{.324} & .194 & 16 & .000 & $-$.208 \\
Jews & 27 & .076 & $-$.067 & 28 & .080 & $-$.072 & 26 & .023 & $-$.059 & 29 & .017 & $-$.022 \\
Other & 32 & .150 & .003 & 32 & .254 & .492 & 16 & .269 & .325 & 31 & $-$.016 & .116 \\
\midrule
\textit{Cross-inst.} & $n$ & $\kappa$ & $\kappa_{\mathrm{pw}}$ & $n$ & $\kappa$ & $\kappa_{\mathrm{pw}}$ & $n$ & $\kappa$ & $\kappa_{\mathrm{pw}}$ & $n$ & $\kappa$ & $\kappa_{\mathrm{pw}}$ \\
\addlinespace[1pt]
All & 6978 & .050 & .045 & 7047 & .034 & .061 & 5963 & .078 & .076 & 6890 & .031 & .071 \\
\addlinespace[1pt]
Muslims & 3598 & .064 & .123 & 3667 & .027 & .057 & 3311 & .041 & .043 & 3557 & .057 & .126 \\
Women & 1237 & $-$.012 & $-$.048 & 1237 & .031 & $-$.035 & 827 & \textbf{.195} & .206 & 1221 & $-$.090 & $-$.036 \\
Black people & 346 & .043 & $-$.014 & 346 & .032 & .191 & 236 & .091 & $-$.027 & 336 & \textbf{.132} & .177 \\
LGBTQ+ & 369 & \textbf{.130} & $-$.033 & 369 & .077 & .101 & 300 & .087 & .011 & 357 & .071 & .108 \\
Immigrants & 887 & .030 & $-$.014 & 887 & .013 & .087 & 764 & .081 & .095 & 880 & .003 & $-$.038 \\
Jews & 467 & .085 & $-$.050 & 467 & .062 & .092 & 463 & .108 & .080 & 467 & .095 & .038 \\
Other & 74 & .128 & .072 & 74 & \textbf{.211} & .268 & 62 & .087 & .013 & 72 & .077 & .140 \\
\bottomrule
\end{tabular}%
}
\end{table}

Agreement varies across target groups, but no group shows consistently higher or
lower agreement across all dimensions and settings. Counterspeech addressing
LGBTQ+ people yields the highest relevance agreement in all three settings,
$\kappa_l = 0.056$ in pointwise evaluation, $\kappa = 0.406$ in same-instance
comparisons, and $\kappa = 0.130$ in cross-instance comparisons, whereas
counterspeech addressing women yields the lowest suitableness agreement in all
three settings, with negative values throughout. For coherence in pointwise
evaluation, $\kappa_l$ ranges from 0.083 for women to 0.178 for the Other group,
and $\rho$ from 0.226 for immigrants to 0.491 for Black people. Same-instance
values vary most, as several groups contain few pairs, for example 16 for
immigrants and 28 for Jews, so these estimates are less reliable.


\section{Conclusion}
\label{sec:conclusion}

This work investigated whether LLMs can reliably evaluate the quality of
counterspeech against online hate speech as a scalable alternative to expert
assessment on the Web. Across 144 model--prompting--dimension configurations per
evaluation setting, spanning pointwise, same-instance pairwise, and
cross-instance pairwise evaluation, agreement with human experts is low: no
configuration exceeds fair agreement, and the best, Qwen 2.5 72B on coherence,
reaches $\kappa_l = 0.191$ in pointwise and $\kappa = 0.245$ in pairwise
evaluation, whereas its rank correlation reaches $\rho = 0.421$. Agreement is
highest for coherence in pointwise evaluation and same-instance comparisons, and
lowest for suitableness, where it remains close to zero in pointwise evaluation,
same-instance comparisons, and cross-instance comparisons.

A Wilcoxon signed-rank test shows no significant difference between pairwise
and pointwise evaluation, $p = 0.073$ for same-instance and $p = 0.188$ for
cross-instance comparisons. No evaluation setting is consistently better,
partly because many models give the same pointwise score to every instance,
which turns all their pairs into ties. When models do vary their scores,
pointwise evaluation can perform clearly better: for same-instance coherence
under zero-shot prompting, Qwen 2.5 72B reaches $\kappa_{\mathrm{pw}} = 0.353$
when its pointwise scores are converted into pairwise labels, compared with
$\kappa = 0.245$ when it judges the pairs directly. A paired cluster bootstrap
shows that this difference of 0.107 is significant, 95\% CI [0.053, 0.163].
Neither few-shot nor CoT prompting consistently improves agreement over
zero-shot, and the highest agreement for each dimension is obtained with
zero-shot or few-shot. Larger models tend to achieve slightly higher
agreement, with significant differences only in pairwise evaluation.

LLMs agree more with one another than with the experts, particularly in
pairwise evaluation. OLMo 2 32B and Llama 3.3 70B reach substantial agreement on
same-instance coherence, $\kappa = 0.764$, whereas OLMo 2 32B's agreement with
the experts in the same configuration is 0.214. The gap is largest for
suitableness, where LLMs agree with one another while their agreement with the
experts is close to zero.

Agreement also varies across hate speech target groups, but no group shows consistently higher or lower agreement, and these differences should not be read as evidence of systematic target-group bias.

In our evaluation, neither larger models nor more
elaborate prompting raised agreement with experts above fair. These findings should be interpreted in light of the evaluation design: pairwise reference
labels were derived from expert pointwise ratings rather than elicited as direct
preferences, and chain-of-thought prompting provided structured evaluation
criteria without generating explicit reasoning. The results therefore do not
establish performance against directly annotated preferences or under
unrestricted chain-of-thought reasoning, and their generalizability beyond the
evaluated dataset, models, and prompts remains to be tested. 

Future work will look further in this direction. Despite the low agreement with the experts, these results could still be leveraged to filter and reduce part of the human work in counterspeech evaluation. The goal would not be to replace human evaluation entirely but to identify those items where human work may be less needed or, vice versa, to highlight those requiring intervention. This may result from an in-depth analysis of the agreement among LLMs: while this is higher than the human--LLM agreement, this is not perfect. Items showing higher disagreement may be worth further investigation. Additionally, since counterspeech effectiveness depends on the specific recipient of such messages, we plan to evaluate the use of personas to prompt, aiming to obtain more tailored assessments.


\bibliographystyle{ACM-Reference-Format}
\bibliography{references}
\clearpage

%

\appendix

\appendix

\section{Prompt and Experimental Details}
\label{app:prompts}

All evaluations consider one quality dimension independently.
The prompts share a common task structure, with the corresponding
CSEval dimension definition, scale or label rule, and prompting
strategy substituted as appropriate.

\subsection{Prompt Templates}
\label{app:prompt-templates}

\paragraph{Pointwise evaluation.}
The LLM receives a hate speech message and one counterspeech
response and assigns a score for a single quality dimension.
Relevance, coherence, and aggressiveness use a 1--5 scale, while
suitableness uses a 1--3 scale. For aggressiveness, lower scores
indicate better quality; for the remaining dimensions, higher scores
indicate better quality. The input structure is:

\begin{quote}
\small\ttfamily
HATE SPEECH (HS): \{hatespeech\}\\
COUNTERSPEECH (CS): \{counterspeech\}
\end{quote}

\paragraph{Pairwise evaluation.}
The LLM compares two counterspeech responses on one quality
dimension and outputs \texttt{A}, \texttt{B}, or \texttt{Tie}.
The counterspeech responses, rather than the hate-speech--counterspeech
pairs as complete units, are the evaluation targets.

For same-instance comparisons, both responses share the same
hate-speech context:

\begin{quote}
\small\ttfamily
HATE SPEECH: \{hatespeech\}\\
COUNTERSPEECH A: \{cs\_a\}\\
COUNTERSPEECH B: \{cs\_b\}
\end{quote}

For cross-instance comparisons, each response is presented with its
corresponding hate-speech context:

\begin{quote}
\small\ttfamily
HATE SPEECH A: \{hate\_a\}\\
COUNTERSPEECH A: \{cs\_a\}\\
HATE SPEECH B: \{hate\_b\}\\
COUNTERSPEECH B: \{cs\_b\}
\end{quote}

\paragraph{Prompting strategies.}
Zero-shot prompting contains the task instructions and the
dimension-specific definition. Pointwise few-shot prompting adds
target-group-specific demonstrations selected from the model-generated
portion of CSEval: five demonstrations covering scores 1--5 for
relevance, coherence, and aggressiveness, and three demonstrations
covering scores 1--3 for suitableness. Pairwise few-shot prompting
uses three demonstrations representing the possible outcomes
\texttt{A}, \texttt{B}, and \texttt{Tie}.

CoT prompting supplements the task definition with fixed
dimension-specific criteria and decomposes each dimension into several
evaluation questions. For example, the
aggressiveness criteria consider whether the counterspeech contains
abusive or offensive language, expresses strong disagreement with the
hateful content, or uses an impolite or disrespectful tone or personal
attacks. These criteria are fixed across models and instances rather
than automatically calibrated from human-annotated examples. Guided
decoding restricts the observable output to a valid score or pairwise label; no free-form reasoning trace is generated.

\section{Statistical Tests}
\label{app:stat-tests}

Table~\ref{tab:stat-tests} summarizes the aggregate tests for prompting
strategy, model size, and pairwise vs.\ converted labels. We report
$\chi^2$ for Friedman tests and $W$ for Wilcoxon signed-rank tests, which
exclude zero differences. These tests report unadjusted $p$-values.
Table~\ref{tab:size-family} reports model-size comparisons within families,
with $p$-values adjusted using the Benjamini--Hochberg procedure within each
setting. Confidence intervals are obtained with the cluster bootstrap described
in Section~\ref{sec:results_discussion}.

\begin{table}[!htbp]
\caption{Aggregate statistical tests. +/$-$: positive/negative differences;
F: Friedman; W: Wilcoxon; ZS/FS: zero-shot/few-shot; PW: pointwise;
CL: converted labels.}
\label{tab:stat-tests}
\centering
\scriptsize
\setlength{\tabcolsep}{3pt}
\begin{tabular*}{\columnwidth}{@{\extracolsep{\fill}}llrrr@{}}
\toprule
Test & Setting & +/$-$ & Statistic & $p$ \\
\midrule
\multicolumn{5}{l}{\textit{Prompting strategy} (48 model--dimension combinations)}\\
F: ZS vs.\ FS vs.\ CoT & PW & -- & $\chi^2=2.93$ & .231 \\
F: ZS vs.\ FS vs.\ CoT & Same & -- & $\chi^2=14.23$ & $<$.001 \\
F: ZS vs.\ FS vs.\ CoT & Cross & -- & $\chi^2=1.79$ & .408 \\
W: FS $-$ ZS & PW & 13/19 & 227 & .500 \\
W: CoT $-$ ZS & PW & 11/20 & 182 & .202 \\
W: FS $-$ ZS & Same & 16/32 & 344 & .012 \\
W: CoT $-$ ZS & Same & 24/23 & 511 & .582 \\
W: FS $-$ ZS & Cross & 19/29 & 447 & .151 \\
W: CoT $-$ ZS & Cross & 22/26 & 584 & .972 \\
\addlinespace
\multicolumn{5}{l}{\textit{Model size} (72 matched small/large pairs)}\\
W: large $-$ small & PW & 32/20 & 486 & .064 \\
W: large $-$ small & Same & 46/26 & 783 & .003 \\
W: large $-$ small & Cross & 59/13 & 338 & $<$.001 \\
\addlinespace
\multicolumn{5}{l}{\textit{Pairwise vs.\ converted labels} (144 configurations)}\\
W: pairwise $-$ CL & Same & 79/65 & 4320 & .073 \\
W: pairwise $-$ CL & Cross & 84/59 & 4495 & .188 \\
\bottomrule
\end{tabular*}
\end{table}

\begin{table}[H]
\caption{Model size within families: configurations in which the larger model
has higher/lower $\kappa$, with Benjamini--Hochberg adjusted $p$-values in
parentheses. PW: pointwise.}
\label{tab:size-family}
\centering
\small
\setlength{\tabcolsep}{10pt}
\begin{tabular}{lccc}
\toprule
Family & PW & Same & Cross \\
\midrule
Llama     & 7/2 (.431)  & 8/4 (.077)  & 9/3 (.014) \\
OLMo 2    & 8/3 (.322)  & 10/2 (.019) & 12/0 (.003) \\
Qwen 2.5  & 7/1 (.407)  & 12/0 (.003) & 10/2 (.014) \\
Mistral   & 6/5 (.431)  & 2/10 (.019) & 9/3 (.677) \\
Mixtral   & 4/8 (.407)  & 9/3 (.077)  & 11/1 (.003) \\
Command R & 0/1 (1.000) & 5/7 (.970)  & 8/4 (.622) \\
\bottomrule
\end{tabular}
\end{table}

\end{document}